\documentclass[11pt,a4paper]{article}
\usepackage{graphicx}
\usepackage[hyperref]{emnlp2018}
\usepackage{times}
\usepackage{latexsym}
\usepackage{amsmath}

\usepackage{amsfonts}
\usepackage{array,booktabs} 
\newcolumntype{P}[1]{>{\centering\arraybackslash}p{#1}}
\usepackage{verbatim}
\usepackage{url}

\aclfinalcopy 

\title{Question-Aware Sentence Gating Networks \\ for Question and Answering}

\author{Minjeong Kim \\
  Dept.of Computer Science\\
  Korea University  \\
  {\tt minjeong1642@gmail.com} \\\And
  David Keetae Park \\
  Dept.of Computer Science\\
  Korea University \\
  {\tt heykeetae@korea.ac.kr} \\\AND
  Hyungjong Noh\\
  NLP Center\\
  NCSOFT\\
  {\tt nohhg0209@ncsoft.com} \\\And
  Yeonsoo Lee\\
  NLP Center\\
  NCSOFT\\
  {\tt yeounsoo@ncsoft.com} \\\And
  Jaegul Choo\\
  Dept. of Comp\\
  Korea University\\
  {\tt jchoo@korea.ac.kr}
  }

\date{}

\begin{document}
\maketitle
\begin{abstract}
Machine comprehension question answering, which finds an answer to the question given a passage, involves high-level reasoning processes of understanding and tracking the relevant contents across various semantic units such as words, phrases, and sentences in a document. This paper proposes the novel question-aware sentence gating networks that directly incorporate the sentence-level information into word-level encoding processes. To this end, our model first learns question-aware sentence representations and then dynamically combines them with word-level representations, resulting in semantically meaningful word representations for QA tasks. Experimental results demonstrate that our approach consistently improves the accuracy over existing baseline approaches on various QA datasets and bears the wide applicability to other neural network-based QA models.  
\end{abstract}

\section{Introduction}
\begin{figure}[t]
\centering
\includegraphics[width=0.5\textwidth]{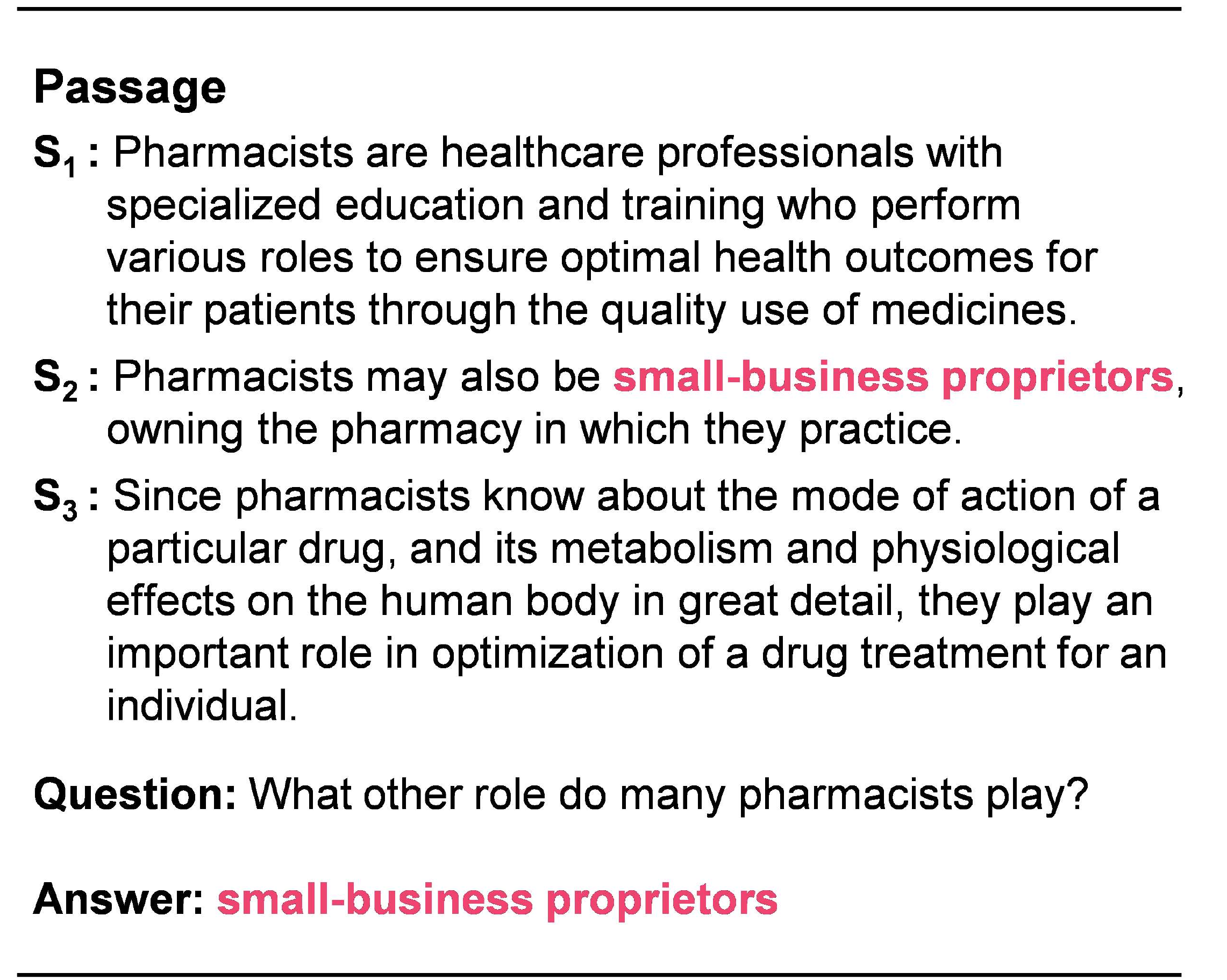}
\caption{An example from SQuAD dataset.}\label{fig:sample_squad}
\end{figure}

A machine comprehension question answering (QA) task has gained significant popularity from research communities. Given a passage and a question, the machine finds the correct answer by understanding the passage using the contextual information from various levels of semantics. In the example shown in Figure~\ref{fig:sample_squad}, the question `What other role do many pharmacists play?' requires the model to properly get the meaning of the first sentence (the first role of the pharmacists), so as to understand that the role presented in the second sentence corresponds to the `other role,' which is needed by the question.
To tackle this machine comprehension, recently, many neural network-based models have been proposed~\cite{seo2017bidirectional, liu2017phase, pan2017memen,huang2017fusionnet,hu2017reinforced}. Most of them are trying to solve this problem by extracting the matching information between the word-level encoding vectors in the passage and those in the question. For example, \citet{xiong2017dynamic} introduces a co-attention encoder that captures the word-level question-passage matching information, and ~\citet{wang2017gated} additionally incorporates self-matching information among words within the passage. 

However, most of the existing approaches still focus on improving word-level information, but not directly utilize the higher-level semantics contained in, say, sentences and paragraphs. Motivated by this, we propose the novel \textit{question-aware sentence gating} networks that directly impose the sentence-level information into individual word encoding vectors. 

Our proposed approach works as follows: (1) \textbf{Sentence encoding}: we first generate initial encoding vectors of individual sentences in the passage and the question. (2) \textbf{Sentence matching}: we further transform them into \textit{question-aware} vectors by considering its degree of matching with the question. (3) \textbf{Sentence gating}: For each word, our gating module determines how much to add the sentence information to it, and the new word-level encoding vector is computed accordingly as a convex combination of its original word vector and its corresponding sentence vector. 
 

Finally, the contribution of our paper includes the following: 
\begin{itemize}
\item We propose the novel question-aware sentence gating networks that impose the sentence-level semantics into individual word encoding vectors in QA tasks. 
\item We extend several existing state-of-the-art models by integrating our sentence gating networks. 
\item We show the effectiveness of our proposed approach by conducting quantitative comparisons using well-known QA datasets and models as well as qualitative analysis of our model. 
\end{itemize}
 
 

\begin{figure}[t]
\centering
\includegraphics[width=0.5\textwidth]{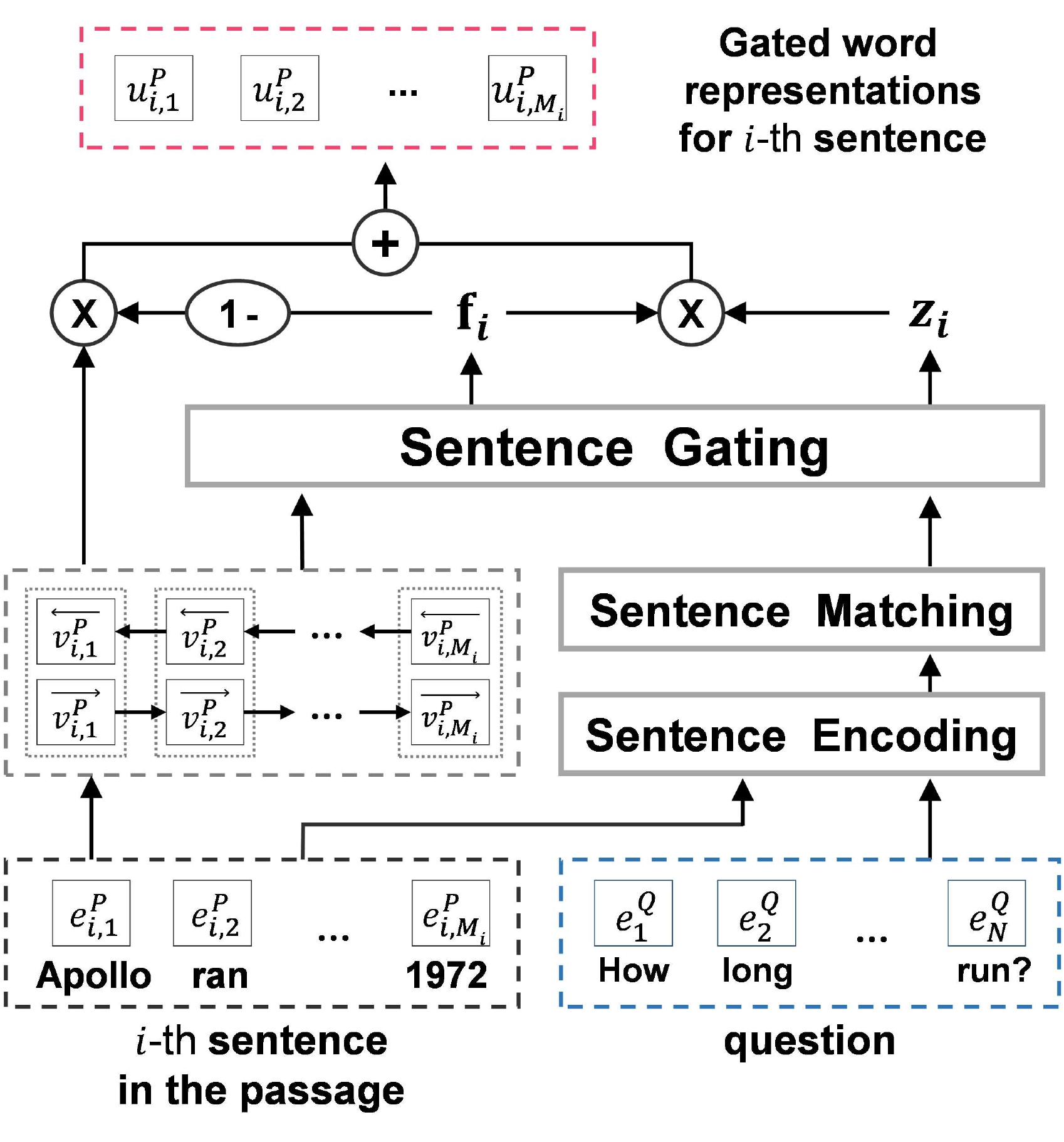}
\caption{Overview of our proposed question-aware sentence gating networks. Inputs of networks are word tokens of the $i$-th sentence in a passage and question tokens. The networks yield the gated word representations for the $i$-th sentence.}\label{fig:SentGate}
\end{figure}

\begin{figure}[t]
\centering
\includegraphics[width=0.5\textwidth]{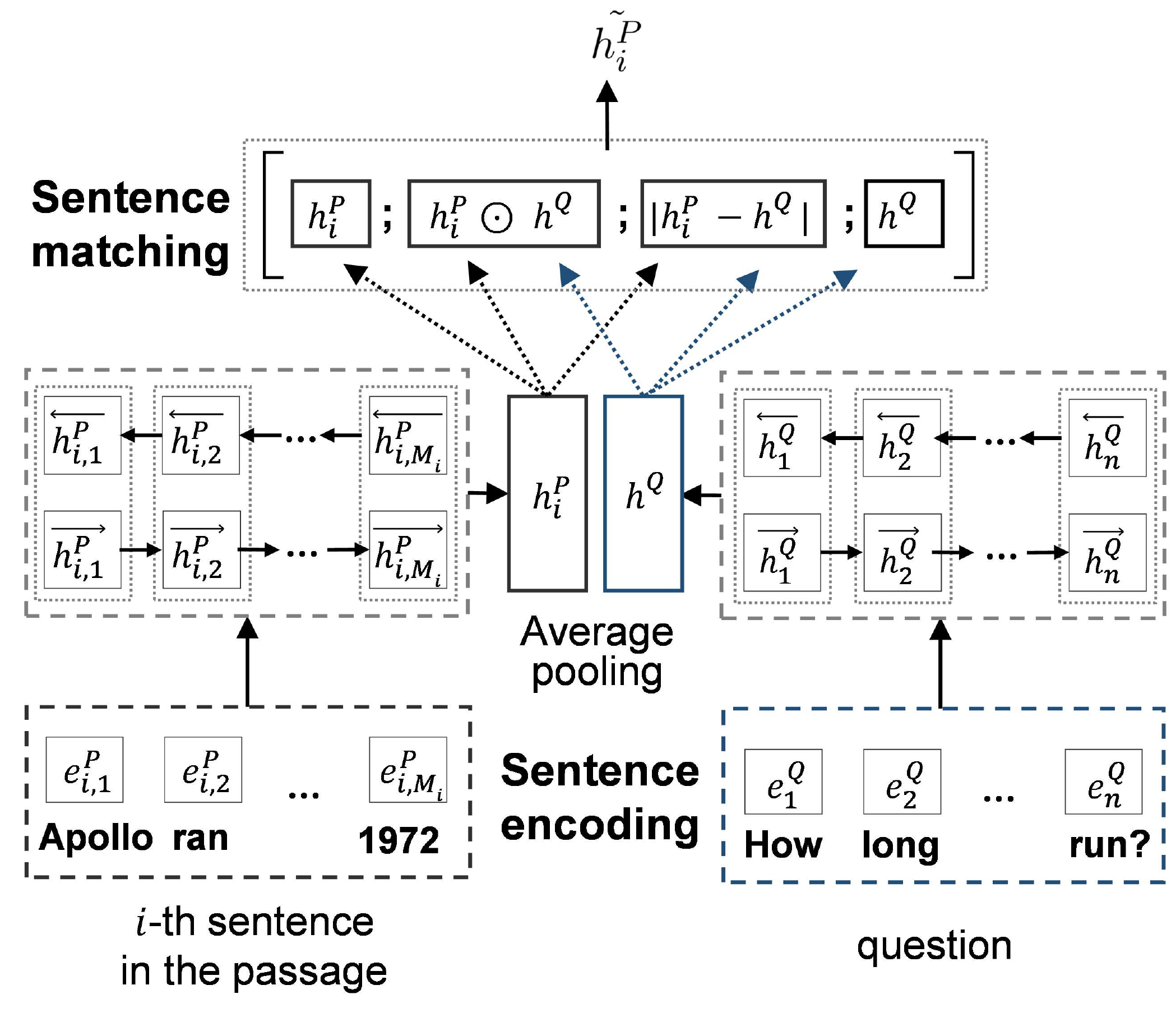}
\caption{A detailed illustration on the sentence encoding and the sentence matching modules. Note that the produced output $\tilde{h_{i}^{P}}$ is passed to the sentence gating module.}\label{fig:SentEncode}
\end{figure}

\section{Problem Setting}
Taking a passage and a question as input, a QA model jointly understands the two and finds out the corresponding answer from the passage. 
The passage is represented as a sequence of sentences, $\ensuremath{\mathbf{P}_s=\left\{ s_{i}^{P}\right\} _{i=1}^{L}}$, where $L$ is the number of sentences in the passage. The $i$-th sentence $s_{i}^{P}$ is further decomposed into a sequence
of words, $\ensuremath{s_{i}^{P}=\left\{ w_{i,j}^{P}\right\} _{j=1}^{M_{i}}}$, where $M_{i}$ denotes the number of words in sentence $s_{i}^{P}$. Alternatively, the passage is represented just as a single sequence of words across multiple sentences, $\mathbf{P}_w=\left\{w_{t}^{P}\right\} _{t=1}^{M}$, where $M=\sum_{i=1}^{L}M_{i}$. 
The (single-sentence) question is similarly denoted as $s^{Q}=\left\{w_{t}^{Q}\right\} _{t=1}^{N}$,
where $N$ denotes the number of words in $s^{Q}$. We assume that the answer always has the form of either a single entity or a phrase existing in the passage.


\section{Question-Aware Sentence Gate}

\label{qa_sentence_gate} In this section, we describe our proposed
question-aware sentence gating networks that directly incorporate
the sentence-level information into the encoding module of individual
words that belong to the corresponding sentence. 
The overview of our proposed method is summarized in Figure~\ref{fig:SentGate}.

\subsection{Sentence Encoding Module}


Given the word sequence of a passage, $\left\{ w_{t}^{P}\right\} _{t=1}^{M}$
and that of a question, $\left\{ w_{t}^{Q}\right\} _{t=1}^{N}$, the
first layer of our model is an embedding layer~\cite{mikolov2013distributed,pennington2014glove},
which represents individual words as pre-trained word embedding vectors
$\left\{ e_{t}^{P}\right\} _{t=1}^{M}$ and $\left\{ e_{t}^{Q}\right\} _{t=1}^{N}$.
Our model then applies bi-directional Gated Recurrent Units (BiGRU)
to each set of embedding vectors in the passage and the question,
respectively, as 
\begin{align*}
h_{t}^{P} & =\textrm{BiGRU}_{PS}\left(h_{t-1}^{P},e_{t}^{P}\right)\\
h_{t}^{Q} & =\textrm{BiGRU}_{Q}\left(h_{t-1}^{Q},e_{t}^{Q}\right),
\end{align*}
where $h_{t}^{P}\in\mathbb{R}^{2d}$ and $h_{t}^{Q}\in\mathbb{R}^{2d}$
are the word-level encoding vector of the $t$-th word in the passage
and the question, respectively, and $d$ is the hidden dimension. 

Afterwards, an average pooling layer is applied to the set of these
word-level encoding vectors per sentence in the passage, i.e., 
\begin{align*}
h_{i}^{P} & =\frac{1}{M_{i}}\sum_{j}^{M_{i}}h_{i,j}^{P}\\
h^{Q} & =\frac{1}{N}\sum_{t}^{N}h_{t}^{Q},
\end{align*}
where $h_{i,j}^{P}\in\mathbb{R}^{2d}$ represents
the $j$-th word of the $i$-th sentence. 

As will be seen in our experimental results (Table~\ref{table:sent_encoding_methods}),
we choose the average pooling approach for sentence encoding since
it outperforms other approaches while using relatively small number
of parameters. 

\subsection{Sentence Matching Module}

\label{sec:sent_match} 

The generated passage and question sentence representations are then
fed into our sentence-matching module to incorporate the question
information into the sentence vectors in the passage.

Inspired by the generic natural language inference approaches~\cite{liu2016learning,conneau2017supervised},
we compute the degree of matching between each passage sentence representation
$h_{i}^{P}$ for $i=1,\,\cdots,\,L$ and the question sentence representation
$h^{Q}$ in a diverse manner. Specifically, we propose the three different
matching schemes~\cite{mou2015natural}: (1) the concatenation of the two representation
vectors $\left(h_{i}^{P};\,h^{Q}\right)\in\mathbb{R}^{4d}$,
(2) the element-wise product $h_{i}^{P}\odot h^{Q}$, where $\odot$ represents
the Hadamard product, and (3) the absolute element-wise difference
$\left|h_{i}^{P}-h^{Q}\right|$. The results of three matching methods
are then concatenated as
\begin{equation}
\tilde{h_{i}^{P}}=\left[h_{i}^{P};\,h^{Q};\,h_{i}^{P}\odot h^{Q};\,\left|h_{i}^{P}-h^{Q}\right|\right],\label{eq:h_tilde_i}
\end{equation}
which works as the question-aware sentence representation $\tilde{h_{i}^{P}}\in\mathbb{R}^{8d}$.
It captures the sentence-level relation between the passage and the
question, which is then fed to the sentence gating module to flow
the sentence-level matching information to the word-level representation.

\subsection{Sentence Gating Module}
\label{sec: sentgate} 

We propose the sentence gating networks that determine how much
the sentence-level matching information $\tilde{h_{i}^{P}}$ should
be incorporated into each word representation towards more semantically
meaningful word representations for QA tasks. 

To this end, we first generate another word-level representations
$v_{t}^{P}$ by using additional BiGRU module given the entire passage
word embedding vectors ${\left\{ e_{t}^{P}\right\} _{t=1}^{M}}$,
i.e., 
\[
v_{t}^{P}=\textrm{BiGRU}_{PW}\left(v_{t-1}^{P},e_{t}^{P}\right).
\]
Note that a newly indexed word-level representation $v_{i,j}^{P}$ corresponds to $v_{t}^{P}$ where the $t$-th word in the passage
is the $j$-th word in the $i$-th sentence. 

Then, our sentence gating module takes the input as the word representation
$v_{i,j}^{P}$ and its corresponding sentence representation $\tilde{h_{i}^{P}}$
(defined in Eq.~\ref{eq:h_tilde_i}) and generates a new word representation
$z_{i,j}\in\mathbb{R}^{2d}$ (fused with the sentence information)
and its dimension-wise gating vector $f_{i,j}\in\mathbb{R}^{2d}$,
respectively, as %
{} 
\begin{align*}
z_{i,j} & =tanh(W_{z}^{(1)}\tilde{h_{i}^{P}}+W_{z}^{(2)}v_{i,j}^{P}+b_{z})\\
f_{i,j} & =\sigma(W_{f}^{(1)}\tilde{h_{i}^{P}}+W_{f}^{(2)}v_{i,j}^{P}+b_{f}),
\end{align*}
where $W_{z}^{(1)},W_{f}^{(1)}\in\mathbb{R}^{2d\times8d},W_{z}^{(2)},W_{f}^{(2)}\in\mathbb{R}^{2d\times2d}$
and $b_{z},b_{f}\in\mathbb{R}^{2d}$ are the trainable parameters,
and $\sigma$ represents an element-wise sigmoid activation function.

As our final output, the  gated word representation
$u_{i,j}^{P}$ is computed as a convex combination of $v_{i,j}^{P}$
and $z_{i,j}$, i.e., 
\begin{equation}
u_{i,j}^{P}=(1-f_{i,j})\odot v_{i,j}^{P}+f_{i,j}\odot z_{i,j},\label{eq:u_ij}
\end{equation}
where $u_{i,j}^{P}\in\mathbb{R}^{2d}$ and $\odot$ represents the
Hadamard product. Each element of $f_{i,j}$ determines the amount
of information in the corresponding dimension of $h_{i}^{P}$ to be
added to the previous-layer word representation $v_{i,j}^{P}$.

This proposed algorithm is widely applicable to the passage and the
question encoding processes in various existing QA models by replacing
the word-level encoding module output with our sentence-gated
word representations $\left\{ u_{t}^{P}\right\} _{t=1}^{M}$, as will
be discussed in Section~\ref{sec: baselines}.

\begin{table}[t]
\begin{center}

\begin{tabular}{cccc}  
\toprule
\textbf{Data} & \textbf{\# Train} & \textbf{\# Validation} & \textbf{\# Test}  \\
\midrule
SQuAD & 87,599 & 10,570& -\\
WDW  & 127,786 & 10,000& 10,000\\
WDW-R & 185,978 & 10,000& 10,000\\
\bottomrule
\end{tabular}
\caption{\label{table:data_statistics}Data Statistics.}
\end{center}
\end{table}

\section{QA Model with Sentence Gate}
\label{sec: baselines}
This section discusses how we extend existing QA models using our proposed sentence gating networks. Specifically, we select two QA models: Match-LSTM with Answer Pointer~\cite{wang2017machine} and Gated-Attention Reader~\cite{dhingra2017gated}. ~\footnote{All the implemented codes will be released upon acceptance.}

\subsection{Match-LSTM with Answer Pointer}

\paragraph{Model Description}
To extend Match-LSTM with Answer Pointer (Match-LSTM), we first generate  gated word representations ${\left\{u^{P}_{t}\right\}_{t=1}^{M}}$ (Eq.~\ref{eq:u_ij}) and encode question word representations ${\left\{h_{t}^{Q}\right\}_{t=1}^{N}}$ using our approach. Next, a bi-directional match-LSTM~\cite{wang2016machine} layer computes attention weights over the entire encoded question word vectors for each encoded passage word, which produces hidden word representations that reflects its relevance to the question. From these vectors, an answer pointer layer selects the final answer span. 

On the other hand, this model originally uses BiGRU to encode the passage and the question words, and the outputs are then fed to the match-LSTM layer. As illustrated in Section~\ref{sec: experiment}, we compare our extended model with the original model. 



\paragraph{Dataset}
We evaluate the extended and the baseline model on Stanford Question Answering Dataset (SQuAD) ~\cite{rajpurkar2016squad}. SQuAD contains more than 100,000 questions on 536 Wikipedia articles. The answer of each question is represented as a span in the passage.

\paragraph{Peripheral Details}
We tokenize all the passages and questions using spaCy~\footnote{https://spacy.io/} library. We use $300$-dimensional pre-trained GloVe word embeddings~\cite{pennington2014glove}~\footnote{We use `glove.840B.300d.zip' downloaded from https://nlp.stanford.edu/projects/glove/} for both passages and questions, and we do not fine-tune the embedding vectors during training. Out-of-vocabulary words are assigned zero vectors, and we do not use character-level embeddings.

We set the batch size as 40 and the hidden vector dimension is set to 150 for all layers. Also, we use a single BiGRU layer for encoding of word tokens. We apply dropout~\cite{srivastava2014dropout} with its rate of 0.2 between adjacent layers. We use ADAMAX ~\cite{kingma2014adam} with the coefficients of $\beta_{1}=0.9$ and $\beta_{2}=0.999$. The initial learning rate is set as 0.002 and multiplied by 0.9 every epoch.

\subsection{Gated-Attention Reader}
\label{sec:GA_Reader}
\paragraph{Model Description} Gated-Attention Reader (GA Reader) is based on a multi-hop architecture and an attention mechanism. We impose our question-aware sentence gating networks across the hops in this model. Specifically, in the $k$-th hop out of $K$ hops, our algorithm first generates gated word vectors $\left(\left\{u_{t}^{P}\right\} _{t=1}^{M}\right)^{(k)}$ and encoded question word vectors $\left(\left\{ h_{t}^{Q}\right\} _{t=1}^{N}\right)^{(k)}$. They are fed to the Gated-Attention module where each word vector in the passage incorporates the matching information from the entire question words. In the last hop, an attention-sum module ~\cite{kadlec2016text} selects the answer based on a probability distribution over the given candidates. 

Similar to Match-LSTM, this model originally uses BiGRU to encode words in the passage and those in the question, and then the gated-attention module takes these outputs as inputs. This model will also be compared with its extension with our proposed module in Section ~\ref{sec: experiment}.

\begin{table}[t]
\begin{center}
\begin{tabular*}{0.49\textwidth}{c @{\extracolsep{\fill}} ccc}
\toprule
 \textbf{Model}    & \textbf{EM} & \textbf{F1} \\
\midrule
\begin{tabular}{c}{}Match-LSTM\\~\cite{wang2017machine}\end{tabular} & $ 0.6474$ & $ 0.7374$\\
\\[-0.9em]
\begin{tabular}{c}{}Match-LSTM\\(Our implementation)\end{tabular} & $0.6661$ & $0.7606$\\
\\[-0.9em]
\begin{tabular}{c}{}\textbf{Match-LSTM}\\\textbf{+ Sentence gate}\end{tabular} & \textbf{0.6814} & \textbf{0.7736} \\
\\[-0.9em]
\bottomrule
\end{tabular*}
\caption{\label{table:squad_performance}Performance comparisons of the question-aware sentence gate over the baseline Match-LSTM on SQuAD dataset.}
\end{center}
\end{table}


\begin{table*}[t]
\begin{center}
\begin{tabular}{ccccc}
\toprule
\textbf{Model }& \textbf{WDW dev} & \textbf{WDW test} & \textbf{WDW-R dev} & \textbf{WDW-R test} \\
\midrule
GA Reader~\cite{dhingra2017gated}& 0.716& \textbf{0.712} &  0.726&0.726\\
\\[-0.9em]
GA Reader (Our implementation)&0.717 &0.710& 0.722& 0.712\\
\\[-0.9em]
\begin{tabular}[c]{@{}c@{}}\textbf{GA Reader}\\ \textbf{+ Sentence gate}\end{tabular} &\textbf{0.722} & \textbf{0.712} & \textbf{0.733}& \textbf{0.727}\\
\\[-0.9em]
\bottomrule
\end{tabular}
\caption{\label{table:cloze_performance}Performance comparisons of our question-aware sentence gate over the baseline GA Reader on WDW and WDW-R datasets.}
\end{center}
\end{table*}

\paragraph{Dataset}
 We conduct the experimental comparison between the extended GA Reader and the original GA model on a cloze-style dataset called Who Did What (WDW)~\cite{onishi2016did}.  It consists of questions with an omitted entity, corresponding to a person name, and the machine is asked to choose the right answer from a set of choices. WDW has two variants, a Strict (WDW) and a Relaxed (WDW-R) version, and we report experimental results on these two datasets.

\begin{table}[t]
\begin{center}

\begin{tabular}{ccc}  
\toprule
 \textbf{Match-LSTM}    & \textbf{EM} & \textbf{F1} \\
\midrule
BiGRU-last & $ 0.6720$ & $ 0.7620$\\
Max pooling  & $ 0.6762$ & $ 0.7698$\\
Inner attention & $ 0.6764$ & $ 0.7697$\\
\textbf{Average pooling}  & \textbf{0.6814} & \textbf{0.7736}\\
\bottomrule
\end{tabular}
\caption{\label{table:sent_encoding_methods}Performance comparisons on different sentence encoding methods.}
\end{center}
\end{table}

\begin{table}[t]
\begin{center}
\begin{tabular}{ccc}  
\toprule
 \textbf{Match-LSTM}    & \textbf{EM} & \textbf{F1} \\
\midrule
Concatenation  & $ 0.6484$ & $ 0.7446$\\
Scalar gate  & $ 0.6738$ & $ 0.7655$\\
Vector gate & $ 0.6724$ & $ 0.7675$\\
Concatenation + Matching & $ 0.6129$ & $ 0.7105$\\
Scalar gate + Matching & $ 0.6747$ & $ 0.7686$\\
\textbf{Vector gate + Matching} & \textbf{0.6814} & \textbf{0.7736}\\
\bottomrule
\end{tabular}
\caption{\label{table:ablations}Performance comparisons over sentence matching and gating methods. \textit{Matching} refers to the sentence matching module illustrated in section~\ref{sec:sent_match}.}
\end{center}
\end{table}

\begin{figure*}[t]
\label{sent_length_analysis}
\centering 
\includegraphics[width=0.95\textwidth]{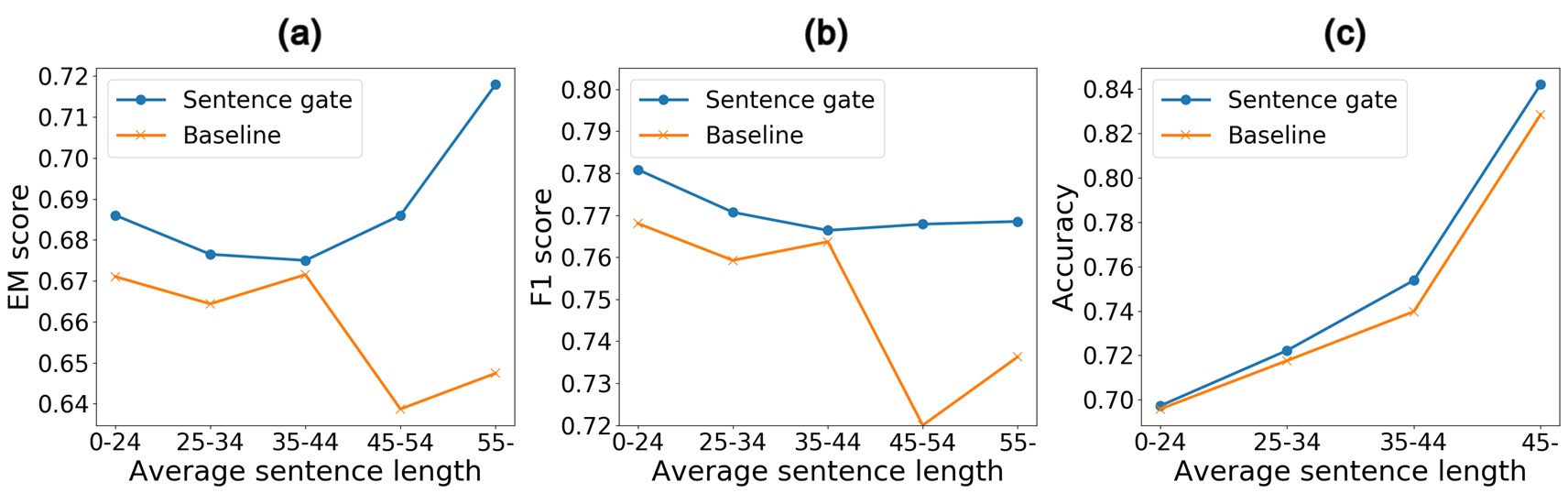}
\vspace*{-3mm}\caption{Performance comparisons depending on various sentence lengths in SQuAD for (a) and (b) and WDW datasets for (c).} \label{fig:sentence_length}
\end{figure*}

\paragraph{Peripheral Details}
 For word embedding, we use the 100-dimensional pre-trained GloVe embedding vectors~\cite{pennington2014glove}, which are fixed during training. Out-of-vocabulary tokens at test time are assigned with distinct random vectors. We train the model with the batch size of 50 and the hidden dimension of 128. We use a single BiGRU layer and adopt a dropout rate of 0.3. The number of hops $K$ is set to 3. We also use the trainable character embedding vectors of 50 dimensions, as in the original model. We also apply the same optimizer and learning rate as in Match-LSTM.

\section{Experiment Results}
\label{sec: experiment}

We analyze our algorithm under two main categories, quantitative and qualitative evaluations. We denote our question-aware sentence gating networks as \textit{Sentence gate} for all the tables and figures presented in this section.

\subsection{Quantitative Evaluations}
In this section, we compare the baseline methods and their extended models we proposed in Section ~\ref{sec: baselines} and also analyze our model structure. Lastly, performance comparisons on various sentence lengths of the passage are presented.

\subsubsection{Model Performance Comparison}
\paragraph{SQuAD}
The performance results on Match-LSTM are reported in Table~\ref{table:squad_performance}. We use two metrics to evaluate the model performance: Exact Match (EM) and F1 score. EM measures the percentage of questions in which the model prediction is exactly matched with the ground truth and F1 measures the harmonic average of the precision and recall between a model prediction and ground truth answer span at character level. Our model outperforms the baseline model by 1.53\% in EM.

\paragraph{Cloze-style}
Table~\ref{table:cloze_performance} shows that the query-aware sentence gate boosts performance on Cloze-style datasets. Our model improves the accuracies on all the benchmarks over reported baseline scores.

\begin{figure*}[t]
\centering 
\includegraphics[width=0.95\textwidth]{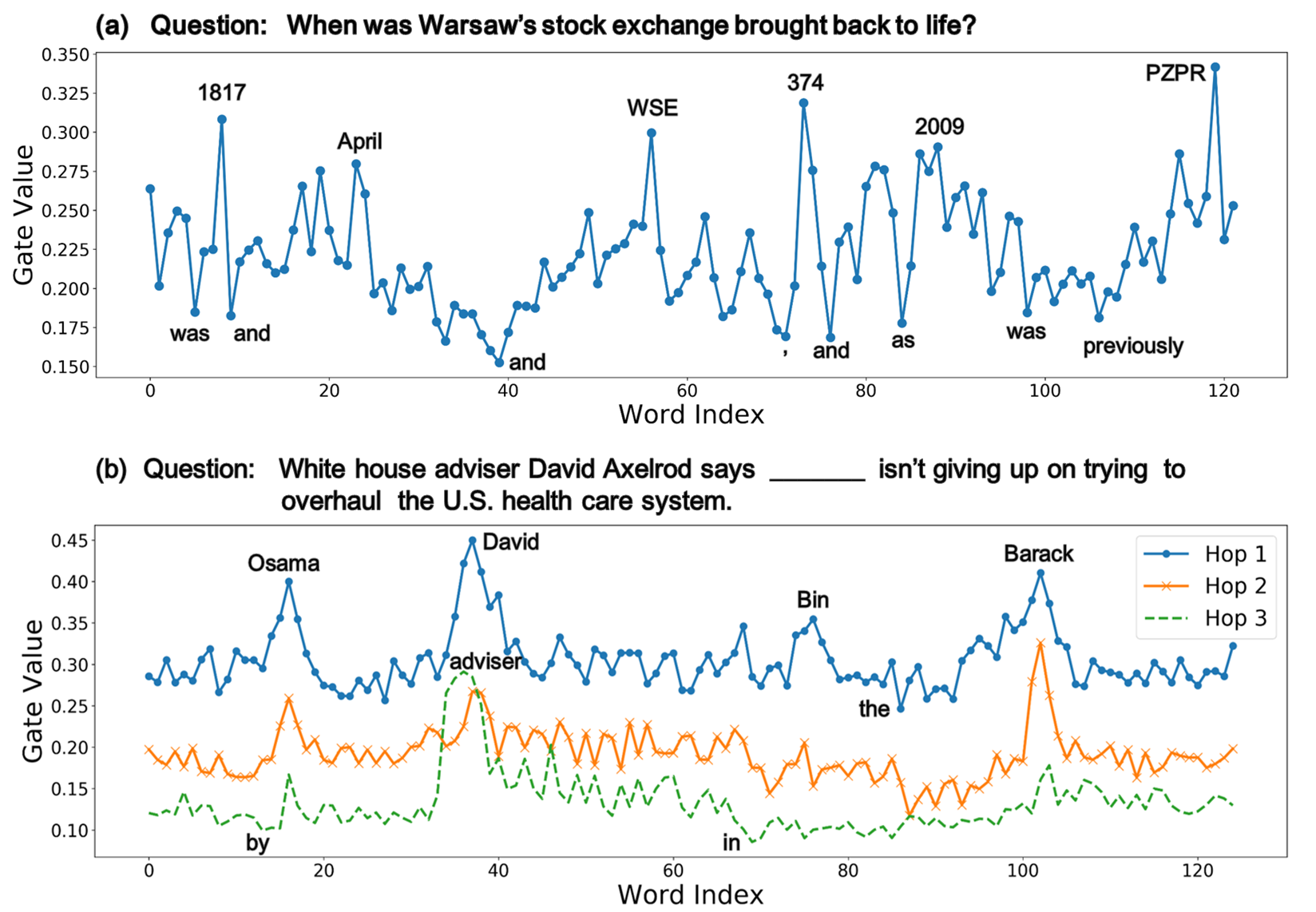}
\vspace*{-3mm}\caption{Word-level visualization of sentence gate values in sample passages.}\label{fig:fi_squad_cloze}
\end{figure*}

\begin{figure}[t]
\label{box_plot}
\centering 
\includegraphics[width=0.465\textwidth]{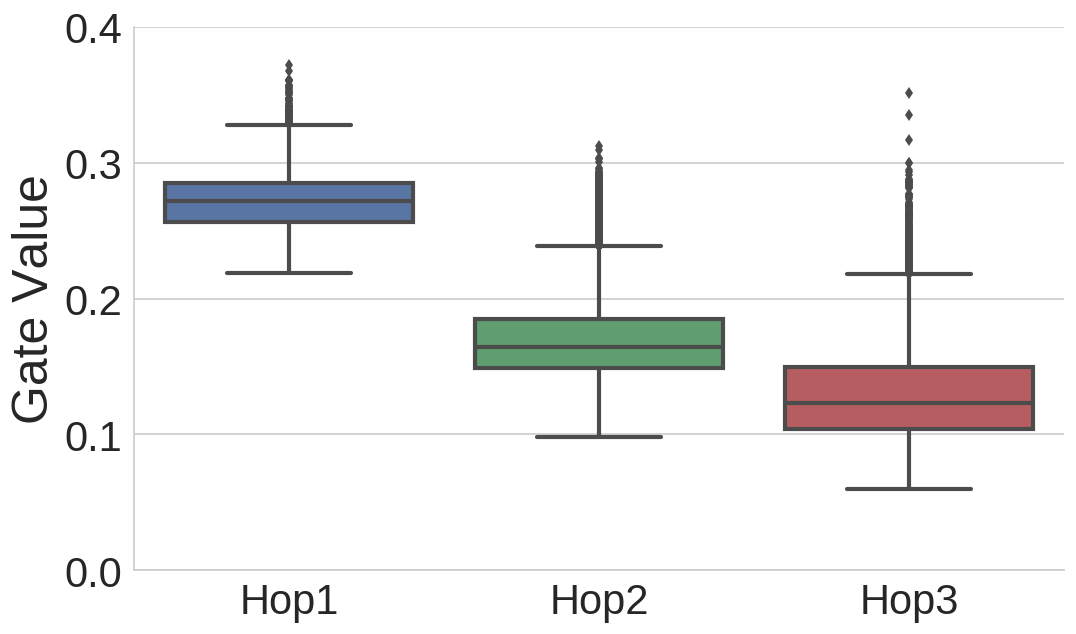}
\caption{Distributions of sentence gate values with respect to different hops}\label{fig:fi_box}
\end{figure}

\subsubsection{Model Structure Analysis}
All the experiments in this sections are performed on the SQuAD dataset with Match-LSTM as the baseline.  
\paragraph{Sentence Encoding Module}
We first compare our average pooling method against the other three sentence encoding methods: 1) BiGRU-last: concatenation of the last hidden states of a BiGRU,
2) Max pooling: taking the maximum value over each dimension of the hidden states,
3) Inner attention: applying an attention mechanism over the hidden states~\cite{liu2016learning,conneau2017supervised}. Table~\ref{table:sent_encoding_methods} shows the results that our approach, i.e., the average pooling, outperforms the other methods.

\paragraph{Sentence Matching and Gating Modules}
We conduct experiments on the effect of the sentence matching and sentence gating modules suggested in Section ~\ref{sec:sent_match} and ~\ref{sec: sentgate}.  We compare the combining methods for word and sentence representations with or without the sentence matching module (referred as \textit{Matching} in Table~\ref{table:ablations}). As for the combining word and sentence vectors, three methods are compared: 1) Concatenation: concatenating sentence and word representations, 2) Scalar gate: applying the same gating weight across dimensions, 3) Vector gate: our proposed method. Also, for the method without the sentence matching module, the output of the average pooling is used for a sentence vector. 

Table~\ref{table:ablations} shows that when it comes to a method for combining word and sentence representation, both the scalar gate and the vector gate outperform the concatenation method by a large margin. Among the two gating mechanisms, the vector gate achieves the highest EM and F1 scores along with the sentence matching module. Also, we can see that the sentence matching module improves the EM and F1 scores of SQuAD dataset, except for the case in conjunction with the concatenation. This indicates that with the proper manner of combining word and sentence, matching information between a passage and question sentence leads to a better model performance.    

\subsubsection{Performance Comparison on Various Sentence Length}
 We observe that our model shows remarkable performance gains for longer sentences compared to the baselines in both datasets. To verify this, we first split passages into five groups based on the average sentence length of each passage and then measure the performance of each group. In Figure~\ref{fig:sentence_length}, (a) and (b) show that our model achieves better performances as the average length of sentence increases in SQuAD dataset. Likewise, our model records a higher accuracy on WDW than the baseline as shown in (c). This implies that our model better tackles a passage which consists of long sentences than word match-based QA models.

\begin{table*}[t]
\begin{center}
\begin{tabular}{cP{6cm}P{6cm}}  

\toprule
 \textbf{Data}    & \textbf{highest} & \textbf{lowest} \\
\midrule
 &Tesla, 2011, Luther, 2016, There, &and, modern, were, being, was, 
 \\
 \textbf{SQuAD} &BSkyB, Kublai, 2015, Goldenson,&very, against, as, largely, highly,  \\
 (Match-LSTM) &Chloroplasts, 2012, Genghis, Newton, & considered, traditional, political,  \\
& TFEU, San, ctenophores, Esiason &extremely, particularly, rich, large
 \\

\addlinespace
\midrule
\addlinespace

&Barack, Robert, Nicolas,  &the, a, of, best, top, final, 
 \\
\textbf{WDW-R} &George, Hugo, Timothy, James, &straight, economic, to, been,  \\
  (GA Reader) &Eric, Alex, Janet, Chudinov,  & last, largest, lost, guaranteed, 
 \\
 & Jacqueline, Shaoxuan&negotiating, first, more, be, all \\

\bottomrule
\end{tabular}
\caption{\label{table:high_and_low_words} Word tokens with the highest and the lowest gate values.}
\end{center}
\end{table*}

\subsection{Qualitative Evaluations}
we also perform in-depth qualitative analyses on the sentence gate value to examine the behavior of our proposed module. 

\subsubsection{Gating Value by Word Tokens}
 Each dimension of the sentence gate value determines the amount of sentence-level information added to the word-level representations.  A higher sentence gate value thus indicates that the corresponding word requires more sentence-level question-passage matching information.

To analyze such behaviors with respect to different words, we sample random passages from SQuAD and WDW-R and plot the sentence gate value of each word token averaged across dimensions. We highlight the words with a relatively high gate value.\footnote{More visualization samples are reported in the supplementary material.} Figures~\ref{fig:fi_squad_cloze}(a) and (b) shows the results for SQuAD and WDW-R, respectively, and each line in (b) corresponds to one of three different hops in the GA Reader. 

Note that word tokens with relatively high gate values are related to the question. For instance, in Figure~\ref{fig:fi_squad_cloze}(a), words related to the question such as \textit{1817}, \textit{April}, \textit{WSE} or \textit{2009} are given high gate values. Likewise, in Figure~\ref{fig:fi_squad_cloze}(b), because the question asks a person name, the names of persons are assigned with high gate values across different hops. On the other hand, we observe that some stop words, e.g., \textit{was}, \textit{and}, and \textit{by}, and adverbs, e.g., \textit{previously} generally take low gate values on both datasets. 
To further justify this observation, we list word tokens with the highest and the lowest gate values in Table~\ref{table:high_and_low_words}. Note that since WDW-R dataset always requires the model to find a correct person name, high gate values are assigned to person names. These results prove our initial hypothesis that the gate values efficiently expose the words to the relevant sentence-level matching information, which is needed to solve given questions.  


\subsubsection{Gate Value by hops}
Once each word token learns sentence-level semantics, it is natural for the model to narrow down the focus on word-level in order to find the exact answer tokens.

To visualize this intuition, we draw a boxplot for gate value distributions over three hops obtained by GA Reader with the WDW-R dataset.  To obtain a single representative gate value for a passage,  we average the  values over dimensions and words. Figure~\ref{fig:fi_box} shows that the average gate value tends to be smaller in later hops. This observation is consistent with our assumption that the model focuses more on word-level information as it approaches to the answer prediction stage.  


\section{Related Works}
In recent years, various tasks on machine comprehension have been attracting considerable attention from the research community. Among the studied tasks are  Automatic Summarization ~\cite{cheng2016neural, nallapati2017summarunner, cao2016attsum}, Machine Translation ~\cite{bahdanau2014neural, kalchbrenner2016neural}, Named Entity Recognition ~\cite{lample2016neural, agerri2016robust} and Question Answering ~\cite{wang2017gated,xiong2017dynamic}.

  
Several studies have been proposing end-to-end neural networks models for QA. Most of the state-of-the-art models in QA utilize the attention mechanism~\cite{bahdanau2014neural} on top of RNN for matching words from a passage to the question or self-matching for better encoding of the passage~\cite{wang2017gated}. For instance, Dynamic Coattention Network~\cite{xiong2017dynamic} employs a co-attentive encoder that iteratively explores the relationship between question and the given context. Bi-Directional Attention Flow networks~\cite{seo2017bidirectional} also utilize the attention mechanism in two different directions to capture the intricate dynamics of questions and contexts. 

While most state-of-the-art models in QA are based on word-level attention mechanism, in Natural Language Inference (NLI) tasks, sentence-level matching methods are popular. For capturing sentence-level semantics, TBCNN~\cite{mou2015natural} introduces the sentence matching module architecture which is widely in use in other studies of the entailment task~\cite{liu2016learning, choi2017unsupervised, chen2017recurrent}. In these studies, the sentence matching module is shown to be effective in extracting relations between sentences. However, this approach is not widely adopted in QA area and to the best of our knowledge only a few works has adopted this method ~\cite{samothrakis2017convolutional}.   

Across the models, Long Short-term Memory ~\cite{hochreiter1997long} networks or Gated Recurrent Unit ~\cite{cho2014learning} are widely used as building blocks for the efficient text encoding. 
The gating mechanism existing in both algorithms allows neural networks models to calculate and select the right amount of necessary information out of two or more distinct modules.
Another application is gating between character and word level embedding vectors for improved representation of hidden states~\cite{yang2016words, miyamoto2016gated}. It has been demonstrated that by a fine-grained gating mechanism, the model benefits from the strengths of both character-level and word-level representations.    

\section{Conclusion}
In this paper, we propose the novel question-aware sentence gating networks. The sentence gating, as our experiments revealed, efficiently and dynamically allows words to take different amount of contextual information  toward more semantically meaningful word representations. Our experiments on two QA datasets, SQuAD and Cloze-style,  demonstrate that the proposed sentence gate improves performance over baseline models with the meaningful behavior of its gate score. Future work involves extending our algorithm to other tasks in machine comprehension.

\clearpage
\bibliography{emnlp2018}
\bibliographystyle{acl_natbib_nourl}
\end{document}